%% file: main.tex
\documentclass[11pt]{article}

\usepackage[final]{acl}

\usepackage{times}
\usepackage{latexsym}
\usepackage{booktabs, multirow}
\usepackage[T1]{fontenc}  
\usepackage[utf8]{inputenc}  
\usepackage{microtype}  
\usepackage{inconsolata}  
\usepackage{graphicx}  
\usepackage{xspace}
\usepackage{cleveref}
\usepackage{subcaption}

\usepackage{tabularx}
\usepackage{bbding}
\usepackage{pifont}
\usepackage{wasysym}
\usepackage{amssymb}
\usepackage{lmodern,babel,adjustbox,booktabs,multirow}
\usepackage[compact]{titlesec}

\usepackage[utf8]{inputenc}

\title{A Review on Subjective Tasks in Farsi: \\From Corpus to Language Model Analysis}
\title{Exploring Subjective Tasks in Farsi:\\ A Survey Analysis and Evaluation of Language Models}

\author{
Donya Rooein$^{1}$ \quad
Flor Miriam Plaza-del-Arco$^{1,2}$ \quad
Debora Nozza$^{1}$ \quad
Dirk Hovy$^{1}$ \\
$^{1}$Bocconi University \quad
$^{2}$Leiden University \\
\texttt{\{donya.rooein, flor.plaza, debora.nozza, dirk.hovy\}@unibocconi.it} \\
}
\begin{document}
\maketitle

\begin{abstract}

Given Farsi's speaker base of over 127 million people and the growing availability of digital text, including more than 1.3 million articles on Wikipedia, it is considered a ``middle-resource'' language. However, this label quickly crumbles when the situation is examined more closely. We focus on three subjective tasks (Sentiment Analysis, Emotion Analysis, and Toxicity Detection) and find significant challenges in data availability and quality, despite the overall increase in data availability.
We review 110 publications on subjective tasks in Farsi and observe a lack of publicly available datasets. Furthermore, existing datasets often lack essential demographic factors, such as age and gender, that are crucial for accurately modeling subjectivity in language. When evaluating prediction models using the few available datasets, the results are highly unstable across both datasets and models. Our findings indicate that the volume of data is insufficient to significantly improve a language's prospects in NLP.

\end{abstract}

\section{Introduction}
Many NLP tasks, like emotion classification, are inherently subjective. There are different valid perspectives on the ``correct'' data labels. How emotions are perceived, for example, differs between the sender and the receiver's subjective interpretations \cite{barz2025understanding}. The same message expressing frustration or sarcasm could be interpreted humorously by one individual yet taken offensively or negatively by another, influenced by their cultural background, personal experiences, or situational context. 

Subjective tasks in NLP, such as emotion analysis, sentiment analysis, and toxic detection, have received increasing attention as they directly impact various societal aspects, including decision making, customer feedback, product evaluation, and the general understanding of social dynamics \cite{nandwani2021review}. These tasks involve assigning texts to specific emotions or sentiments that best reflect the author's mental or emotional state \cite{tao2020toward}. Recent surveys in emotion and sentiment analysis \cite{murthy2021review,kusal2022review, SINGHTOMAR202394, hung2023beyond,venkit-etal-2023-sentiment,al2024challenges,plaza-del-arco-etal-2024-emotion} have primarily focused on identifying available datasets, reviewing models, exploring detection techniques across various modalities (e.g., visual, vocal, textual), and discussing applications. These studies focus on English and do not consider other languages, such as Farsi\footnote{Also known as Persian.}.

Language technologies play a crucial role in promoting multilingualism and preserving linguistic diversity worldwide. However, many languages still face challenges in resource availability, particularly for subjective tasks, despite having substantial digital resources and peer-reviewed research. This is the case for Farsi, which has over 1.3 million Wikipedia articles\footnote{\url{https://en.wikipedia.org/wiki/Persian_Wikipedia}} and has been classified by \citet{joshi-etal-2020-state} as a language with a strong web presence but insufficient efforts in labeled data collection, ranking just below high-resource languages.
Despite these resources, \textbf{research on subjective tasks in Farsi remains notably scarce}, making it a low-resource language in this domain.

While a few survey studies in Farsi focus on sentiment analysis and discuss resource limitations and methodological developments \cite{rajabi2021survey, asgarnezhad2021persian, Borowczyk+2023+1+24}, to the best of our knowledge, no existing work provides a comprehensive survey of multiple subjective tasks in Farsi. This study fills that gap by evaluating encoder-only models and some LLMs across three key tasks: emotion analysis (EA), sentiment analysis (SA), and toxic detection (TD). This gap is particularly concerning in the era of LLMs, where these systems are not only widely accessible but also increasingly used for subjective discussions \cite{ouyang-etal-2023-shifted}. It is essential to evaluate their ability to understand and process sentiments and emotions in Farsi, as well as assess their handling of toxicity to ensure safe and responsible interactions. The absence of research in this area highlights the urgent need for a focused exploration, ensuring that Farsi, like other languages, benefits from advancements in subjective NLP.




We collect relevant studies from publications drawn primarily from ACL Anthology\footnote{\url{https://aclanthology.org/anthology+abstracts.bib}}, and complemented by additional searches on Google Scholar\footnote{\url{https://scholar.google.com/}}. We report the available dataset for each task, including important metadata such as dataset size, labels, and source.
Additionally, we use various language models on selected datasets to assess their capabilities in performing these subjective tasks in Farsi.

We present the following key contributions:

\begin{itemize}
    \item A detailed survey of publications, datasets, and resources specific to the three subjective tasks in Farsi: sentiment analysis, emotion analysis, and toxicity detection.
    \item An experimental evaluation of encoder-only multilingual models and open-source LLMs on these tasks in Farsi.
    \item An analysis of the impact of text translation as a potential solution to address low-resource challenges.
\end{itemize}

\section{Background}
Subjective tasks such as EA, SA, and TD often pose unique challenges due to their reliance on context, cultural nuances, and linguistic features. The EA involves classifying emotions expressed in a text (e.g., joy, sadness, anger) \cite{alm-etal-2005-emotions}. For instance, recognizing the nuanced difference between Farsi expressions like ``delash gereft'' literally his/her heart became tight'') conveying sadness, versus ``delshooreh dārad'' literally ``he/she has a salty heart'' depicting anxiety, requires deep cultural and contextual understanding compared to relatively straightforward English expressions like ``feeling sad'' or ``feeling anxious''. 
The SA consists of determining the sentiment polarity of a text, typically positive, negative, or neutral \cite{wilson2005recognizing}. For example, the Persian expression ``jāye to khālie'' literally ``your place is empty'' carries a positive sentiment, often implying affection, inclusion, and the speaker expresses a desire for the listener’s presence. However, translated directly into English, it may suggest loneliness, absence, or even negativity. Such examples underscore the importance of accurately capturing sentiment, which requires sensitivity to cultural context and linguistic nuances. Toxicity detection consists of identifying language or content considered harmful, offensive, abusive, hateful, or otherwise inappropriate \cite{pavlopoulos-etal-2020-toxicity}. The interpretation of what constitutes toxic content often varies significantly based on cultural and societal norms. For example, the phrase ``aghlet kame'' means ``you're not very smart'' in Farsi, might be considered mildly humorous among close friends, but is perceived as offensive in formal or public contexts.

\section{A Survey on NLP Studies Covering Subjective Tasks in Farsi}
To identify relevant papers with resources related to
EA, SA, TD tasks in Farsi, we design a structured search query consisting of three main components: \texttt{<Task>}, \texttt{<Dataset>}, and \texttt{<Language>}\footnote{All searches are updated by March 2025.}. The \texttt{<Task>} component includes the three NLP tasks we explore: the EA, SA, and TD.
To ensure a comprehensive selection of studies on these tasks, we focus on identifying papers whose titles or abstracts include keywords associated with each task.
For the EA task, our query includes the terms ``emotion classification'', ``emotion detection'', ``emotion recognition'', ``emotion analysis'', and ``emotion prediction''. For the SA task, we incorporate the following keywords: ``polarity classification'', ``sentiment classification'', and ``sentiment analysis''. Lastly, for the TD task, we use terms including ``hate speech detection'', ``offensive language detection'', ``hate speech classification'', ``offensive language classification'', ``toxic detection'', and ``toxic classification''.
The \texttt{<Dataset>} component includes related terms, i.e., \textit{``data set,'' ``dataset,'' ``corpus''}, and \textit{``corpora''}. Finally, the \texttt{<Language>} component focuses explicitly on language-related terms, namely, \textit{``Farsi''} and \textit{``Persian''}. Our query variations derive from 5 keywords associated with the EA, 3 to the SA, and 6 to the TD tasks (the total of 14 keywords), combined with 4 dataset formulation strategies and two for the language, yielding a final calculation of 112 unique phrase searches. To further expand our search, we also collect publications using only \texttt{<Task>} and \texttt{<Language>}, adding 28 additional search phrases. In total, we executed 140 unique phrase searches. We find 12 unique papers from the ACL Anthology: eight focused on SA, four addressed EA, and none focused on the TD task. This absence indicates the lack of research and publicly available datasets on Farsi toxicity detection in the ACL Anthology. To expand our search results, we also use Google Scholar.  Google Scholar lists papers from different research databases; however, it is difficult to verify all the returned sources. We use the SerpApi\footnote{\url{https://serpapi.com/}} library to retrieve papers from Google Scholar. To limit the search results from this engine, we configure the SerpApi to only return the top 10 relevant papers for a given search keyword. This limitation allows us to verify their publishers manually. This search strategy adds 98 more papers which 40 from arXiv\footnote{\url{https://arxiv.org/}}, 16 from IEEE\footnote{\url{https://www.ieee.org/}}, 12 from Springer\footnote{\url{https://www.springer.com/}}, and 30 from other publishers. 

\begin{figure}
    \centering
    \includegraphics[width=\linewidth]{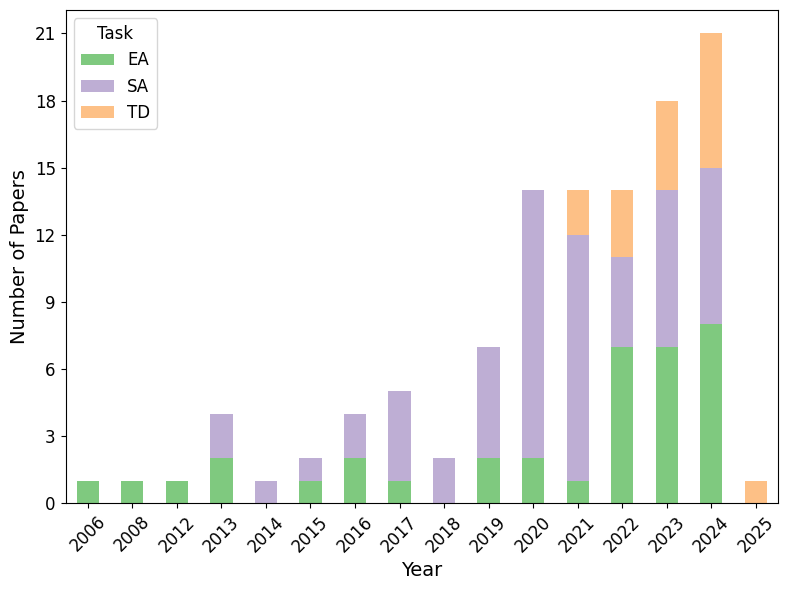}
    \caption{Distribution of papers considered in our
survey by year and tasks (EA: Emotion Analysis, SA: Sentiment Analysis, and TD: Toxicity Detection).}
    \label{fig:trends}
\end{figure}

Thus, we have a total of 110 papers\footnote{The list of the reviewed papers is available at \url{https://github.com/donya-rooein/subjective_tasks_farsi/}}(only 11\% from ACL Anthology), with 36 papers for EA, 58 papers for SA, and 16 papers for TD\footnote{Three of these papers are in the Farsi language and were published at local conferences within Iran.}. Figure~\ref{fig:trends} shows the statistics of the collected papers published from 2006 to 2025. The SA task represents the largest share at 52.7\% (58 out of 110) of all papers, followed by EA at 32.7\% (36 out of 110). The EA task began among the non-NLP community in 2006, focusing on EA through speech. The number of publications remained low in the early years; however, by 2024, the EA in Farsi had increased to 8 papers incorporating text-based modalities. The TD task, which did not appear until 2021, already accounts for nearly 14.5\% (16 out of 110) of papers by 2025, indicating that TD is becoming an increasingly important area of research in NLP Farsi.


\subsection{Annotation Criteria}
After identifying relevant papers, we conduct a manual annotation to summarize and categorize the papers based on consistent criteria. The motivation here is to identify publicly available datasets in Farsi for each task. We adopt the annotation framework proposed by \citet{plaza-del-arco-etal-2024-emotion}, which suggests surveying EA datasets based on five key aspects: annotation framework, language, multimodal, content source, and dataset size. We expand this framework to all the considered subjective tasks and include additional details: lexicon, the type of classification task (e.g., binary, multiclass, or multilabel), and, specifically for studies involving dataset creation, whether the demographics of annotators are explicitly considered. We also include information on the availability of datasets used in each paper. 



Our annotation results reveal several trends. For the data modalities, most works (86.4\%) are only text-based, a few (4.5\%) of research combine text with speech, and 8.2\% studies focus on only speech datasets. In addition, only one paper (0.9\%) uses acoustic and visual data. We identify three categories of papers based on our review collection: (I) papers without datasets, (II) papers with datasets that are not publicly available, and (III) papers with publicly available datasets. We identify 17 out of 36 papers on EA as dataset papers, while only 7 of them provide publicly available datasets. In particular, 4 of these 7 datasets are from the ACL Anthology. 
For SA, we find 33 dataset papers and only 5 available datasets (3 from ACL Anthology). Finally, TD has 14 dataset papers with 3 publicly available datasets.


The datasets used in the reviewed papers are from social media platforms, e-commerce websites, and specialized corpora. The most frequently used sources for social media for all tasks are X\footnote{\url{https://x.com/}} (previously Twitter) and Instagram\footnote{\url{https://www.instagram.com/}}. The e-commerce source is mostly Digikala\footnote{\url{https://www.digikala.com/}}, Iran's largest online retail platform, which contains extensive user-generated product reviews that are valuable for sentiment analysis. Additional sources include datasets from Booking.ir\footnote{\url{https://www.booking.ir/}}, a popular platform for hotel reviews, movie review comments\footnote{From \url{https://cinematicket.org/}}. In some cases, authors use specialized resources such as radio plays or collect datasets from surveys of specific populations. 

\subsection{Datasets} 
Our survey analysis identified 15 publicly available datasets across all tasks (7 for EA, 5 for SA, and 3 for TD). \Cref{tab:datasets} presents a list of publicly available datasets along with detailed information on their names, label sources, data sources, sizes, and modalities.

\begin{table*}
  \centering
  \begin{adjustbox}{max width=\textwidth}
  \begin{tabular}{l|l|l|c|c|l|c}
    \toprule
    \textbf{Task} & \textbf{Dataset} &  \textbf{Mult.} & \textbf{Labels}  & \textbf{Source} & \textbf{Size} & \textbf{Included}\\
    \midrule
    EA & Shemo  & T, S &  E + [neutral] & radio plays & 3,000 & - \\
    EA & ShortPersianEmo & T & [happiness, sadness, anger,
fear, other] & e-commerce & 5,472 & - \\
EA & SAT  & T & E + [anxious, ashamed, disappointed, envious, guilty, insecure, loving, jealous] & chatbot conv. & 5,600 & -\\
EA & ArmanEmo  & T & E - [disguss] + [hate, other] & social media & 7,000 & \checkmark\\
EA & LetHerLearn  & T &  E + [other] & social media & 7,600 & \checkmark\\
EA & LearnArmanEmo & T &  E + [other] & social media & 14,880 & -\\
EA & EmoPars  & T &  E - [disgust] + [hatred] & social media & 30,000 &\checkmark\\
\midrule    
SA & SentiPers & T & [$-2$, $-1$, $0$, $+1$, $+2$] & ecommerce & 15,683 & \checkmark \\
SA &  PersEng & T &  [negative, neutral, positive]  & social media & 3,640 & - \\
SA & Persian Digikala & T &  [negative, neutral, positive]  & e-commerce & 34,465 & -\\
SA  & Pars-ABSA & T &[negative, neutral, positive]  & e-commerce& 10,002  & \checkmark\\
SA & MirasOpinion & T & [$-1$, $0$, $+1$] & e-commerce & 93,868  & \checkmark\\
    \midrule
TD & Phate & T & [hateful (violence, hate, vulgar), normal]& social media &7,056  & \checkmark\\
TD & PHICAD & T & [hate/offense, obscene, spam, none] & social media & 300,000 & \checkmark \\
TD & Pars-OFF  & T & [offensive, not-offensive] & social media & 8,334 & \checkmark\\
     \bottomrule
  \end{tabular}
  \end{adjustbox}
  \caption{Overview of publicly available and private datasets used for subjective tasks in Farsi. \textbf{Task} presents Emotion Analysis (EA), Sentiment Analysis (SA), and Toxicity Detection (TD). The columns provide details on the dataset name if provided (\textbf{Dataset}), which content modality that dataset uses (\textbf{Mult.}), annotation labels (\textbf{Labels}), source of the data (\textbf{Source}), the dataset size (\textbf{Size}), and if they are included in our experiments (\textbf{Included}). [E] Ekman framework. [T] Text and [S] Speech.}
  \label{tab:datasets}
\end{table*}

\textbf{EA datasets:} We identify seven datasets for EA. The \textbf{Shemo} \cite{9721504} dataset is derived from radio plays and annotates five primary emotions, i.e., anger, fear, happiness, sadness, and surprise along with a neutral category, comprising 3,000 samples. This dataset is the only dataset with both text and speech modality, and the rest of the datasets are text-only.
\textbf{ShortPersianEmo}~\cite{sadeghi2021automatic} is from comments on the Digikala website, an e-commerce platform in Iran. The \textbf{SAT}~\cite{elahimanesh2023words} dataset originates from chatbot conversations and distinguishes a broader spectrum of emotions (happy, angry, anxious, ashamed, disappointed, disgusted, envious, guilty, insecure, loving, sad, and jealous) across 5,600 samples. The SAT dataset also includes the demographic information (age and gender) of participants. \textbf{ArmanEmo}~\cite{mirzaee2022armanemo} and \textbf{LetHerLearn}~\cite{hussiny-ovrelid-2023-emotion}, \textbf{EmoPars}~\cite{sabri-etal-2021-emopars-collection} consist of tweets annotated with common emotions such as anger, fear, sadness, happiness, and either wonder or surprise. In particular, EmoPars is annotated by a multilabel annotation approach, assigning a numerical value between 0 and 5 to each emotion (anger, fear, happiness, hatred, sadness, and wonder).
None of these datasets fully adhere to well-known frameworks for emotion analysis such as Ekman's framework \cite{ekman1999basic} which includes anger, fear, sadness, joy, disgust, and surprise or Plutchik’s model \cite{Plutchik}, which encompasses eight primary emotions: anger, anticipation, disgust, fear, joy, sadness, surprise, and trust. \textbf{LearnArmanEmo}~\cite{hussiny-etal-2024-persianemo} combines ArmanEmo and LetHerLearn by unifying their labels based on Ekman's framework. In this unified approach, the label ``happiness'' is used instead of ``joy'', and an ``other'' category is added to capture emotions outside the defined set.

\begin{table*}[ht]
\centering
\small
\begin{tabular}{c|p{12cm}}
\textbf{Task} & \textbf{Prompt} \\
\toprule
EA & Given a text, identify the main emotion expressed. You have to pick one of the following emotions: [List of dataset labels]. Text: \{input\} Only answer with the emotion and omit explanations. Emotion: \{output\} \\
\midrule
SA & Given a text, identify the sentiment expressed. You have to pick one of the following sentiments: [List of dataset labels]. Text: \{input\} Only answer with the sentiment and omit explanations. Sentiment: \{output\} \\
\midrule
TD & Does the following text contain [hate speech/ offensive language]? Only answer with yes or no. Text: \{input\}, Hate speech: \{output\} \\
\bottomrule
\end{tabular}
\caption{Prompt templates for Emotion Analysis (EA), Sentiment Analysis (SA), and Toxicity Detection (TD) tasks.}
\label{tab:prompts}
\end{table*}

\textbf{SA datasets:} Pars-ABSA \cite{shangipour-ataei-etal-2022-pars}, Persian Digikala \cite{kobari2023weighted}, and Persian-English code-mixed datasets \cite{sabri2021sentiment} categorize sentiment of Farsi sentences into positive, negative, and neutral labels. In particular Persian-English code-mixed dataset provides 3,640 labeled tweets, making it one of the few resources addressing sentiment in code-mixed Persian-English text. \textbf{SentiPers} \cite{hosseini2018sentipers} contains 15,683 Digikala reviews annotated on a five-point scale ranging from $-2$ to $+2$. \textbf{MirasOpinion} is the largest available dataset collected from Digikala for SA in Farsi language with 93,868 samples. They label each sentence by using a
Telegram\footnote{\url{https://web.telegram.org/}} bot to several users. They ask them to label the represented document as positive, negative, or neutral. 

\textbf{TD datasets:} We find three datasets, each exclusively in text. \textbf{Phate} \cite{Delbari_Moosavi_Pilehvar_2024} contains tweets that distinguishes between hateful content (with subcategories of violence, hate, and vulgar) and normal content, comprising 7,056 samples. The \textbf{PHICAD} \cite{davardoust_2024_DarkSide} dataset is significantly larger, containing 300,000 samples, and labels content into hate/offense, obscene, spam, or none, also sourced from comments on the Instagram platform. Lastly, \textbf{Pars-OFF} \cite{9936700} focuses on a binary classification of offensive versus not-offensive content with 8,334 samples of tweets.

These datasets, while valuable for advancing Farsi subjective analysis tasks, face several limitations. Many of them exhibit a narrow focus in terms of data sources, mostly based on tweets and comments on the Digikala platform, which may limit the generalizability of models trained on them to other contexts. 
Moreover, they also suffer from the lack of demographic information. Only two datasets of EA (Shemo and SAT) provide the demographic factors (e.g., gender in Shemo and age and gender for SAT). Only authors of three datasets \cite{yazdani2022persian} provide detailed documentation on how annotations were conducted, whether multiple annotators were used, or what inter-annotator agreement was achieved. Without such information, it is difficult to assess the reliability of the labels used to train or evaluate models.

Evaluating these datasets using LLMs may help address some of these shortcomings. \citet{abaskohi-etal-2024-benchmarking} shows the low performance of GPT3.5 and GPT4\footnote{\url{https://openai.com/}} on the emotion recognition task using only the ArmanEmo dataset. In the following section we extend these evaluations by using various open-source models and datasets.

\section{Evaluation Setting}
\subsection{Data}
\label{sec:datasets}
To measure the performance of language models in subjective tasks in Farsi, we select three datasets for each subjective task. For EA, we use ArmanEmo, LetHerLearn, and EmoPars. Since EmoPars contains multilabel emotions, we filter the dataset to include only samples in which one emotion has a non-zero value while all others are zero. With this approach, we reduce the EmoPars dataset size to 5,226 samples. We exclude the Shemo dataset because it relies on speech data, and the transcriptions alone do not adequately capture the nuances of emotion. We also excluded the SAT dataset due to its excessive number of labels, which could negatively impact the performance of language models. Finally, we eliminate the LearnArmanEmo dataset since it is derived from the LetherLearn and ArmanEmo datasets.
For SA, we use ParsABSA, SentiPers, and a subsample of MirasOpinion. Since MirasOpinion is a very large dataset, we only test our language models based on 30k randomly selected samples. We exclude the Persian-English code-mixed dataset due to its limited size and its primary focus on code-mixed vocabulary in Persian.
For the TD tasks, we use all the available datasets presented in \Cref{tab:datasets}. Given that the PHICAD dataset is extensive, with 300,000 samples, we experiment on a subsample provided by \citet{davardoust_2024_DarkSide}\footnote{Part 1 available at \url{https://github.com/davardoust/PHICAD}}of the dataset with 131,959 instances.

\subsection{Models}
\subsubsection{Open Source Decoder-only Models}
From the family of decoder-only LLMs, we select three instruction-tuned versions of popular open-source models which are Meta-Llama-3-8B-Instruct \cite{dubey2024llama} Mixtral-8x7B-Instruct-v0.1 \cite{jiang2024mixtral}, and Qwen2-7B-Instruct \cite{yang2024qwen2}. For each task, we use a zero-shot approach to detect the relevant labels of emotions for EA, sentiments for SA, and hate speech/offensiveness for TD. We use two different prompting strategies on a subset of EA and SA datasets (see \Cref{appx:promting}), then we use the following prompt template that yielded the best performance across these datasets. For TD, we exclusively use one prompt that is introduced by \citet{Delbari_Moosavi_Pilehvar_2024}. We summarize the list of prompts in \Cref{tab:prompts}. For the EA and SA template, we ask the model to identify the main emotion and sentiment expressed in the text, selecting from a predefined list of dataset-specific labels.

\begin{table}[htbp]
\centering
\begin{adjustbox}{max width=\columnwidth}
\begin{tabular}{llclccc}
\toprule
\textbf{Task} & \textbf{Model} & \textbf{Lang.} & \multicolumn{2}{c}{\textbf{Template}} & \textbf{Avg. F1} \\
\cmidrule(lr){4-5}
             &                &               & (I) & (II) & \\
\midrule
\multirow{6}{*}{\textbf{EA}} 
& Llama3-8B     & FA            & 0.19 & 0.19 & 0.19 \\
&            & EN            & 0.18 & 0.20 & 0.19 \\
& Mixtral-7B    & FA            & 0.20 & 0.19 & 0.19 \\
&            & EN            & 0.20 & 0.19 & 0.19 \\
& Qwen2-7B       & FA            & 0.19 & 0.20 & 0.19 \\
&            & EN            & 0.20 & 0.17 & 0.18 \\
\midrule
\multirow{6}{*}{\textbf{SA}} 
& Llama3-8B     & FA            & 0.46 & 0.64 & 0.55 \\
&            & EN            & 0.46 & 0.48 & 0.47 \\
& Mixtral-7B    & FA            & 0.50 & 0.77 & 0.63 \\
&            & EN            & 0.48 & 0.54 & 0.51 \\
& Qwen2-7B       & FA            & 0.48 & 0.36 & 0.42 \\
&            & EN            & 0.48 & 0.46 & 0.47 \\
\bottomrule
\end{tabular}
\end{adjustbox}
\caption{The performances of LLMs in macro average F1 scores for two prompting templates on the EA task for the EmoPars and SA for the MirasOpinion are reported. We use Farsi (FA) and English (EN) versions of datasets (Lang.). The EN version is translated by the NLBB model.}
\label{tab:combined-macro-avg-comparison}
\end{table}

\subsubsection{Data Translation Experiments}
\citet{etxaniz-etal-2024-multilingual} suggest that translating non-English datasets to English can enhance the performance of multilingual LLMs. We adopt this strategy by translating our datasets to assess their impact on model results. Since multiple machine translation systems are available, we first translated a subsample of 100 Farsi sentences using Google Translate\footnote{\url{https://translate.google.com/}}, the NLBB model \cite{costa2022no}, and GPT-4o. After manual evaluation, we found that Google Translate produced the lowest-quality translations. Both NLBB and GPT-4o provided acceptable results, though they still exhibited issues such as literal translations, mistranslations, and omissions. Ultimately, we chose to use NLBB due to its open-source availability.


\subsubsection{Encoder-only Models}
For encoder-only architectures, we use classical fine-tuning approaches using the XLM-RoBERTa model \cite{conneau-etal-2020-unsupervised}. XLM-RoBERTa is a multilingual transformer-based language model pre-trained on data from over 100 different languages. We fine-tune XLM-RoBERTa on nine datasets covering EA, SA, and TD tasks. Fine-tuning is performed by adding a classification head on top of the model’s final hidden representations and optimizing it using a cross-entropy loss function.

\section{Results}
\input{tables/results_models_tasks.tex}





In this section, we present the outcomes of our experiments, detailing the evaluation of prompt selection, LLMs' performances on the datasets in Farsi and their translation in English, and the fine-tuning approach. 

\subsection{Experiment 1: Prompt Variations and Data Translation}

Prompt variations, even the smallest of perturbations such as adding a space at the end of a prompt, can affect the LLM's output \cite{salinas-morstatter-2024-butterfly}. In this regard, we include two prompting strategies: the first involves directly asking the LLM to identify the subjective label of a given text, while the second includes the data source of the text as part of the prompt. For EA, we use a subsample from the EmoPars dataset, and for SA, we select the subsample of the MirasOpinion dataset. We choose these two publicly available datasets, because they are from the ACL Anthology and they have the largest sample sizes, with sample sizes of 5,226 for EmoPars and 30k for MirasOpinion. We evaluate two distinct prompt templates, as described in \Cref{appx:promting}, on these sub-samples.

\Cref{tab:combined-macro-avg-comparison} shows the performance of selected LLMs in EA and SA tasks over selected sub-samples in Farsi and English. The results of EA exhibit low F1-scores (between 0.18–0.20) across all models and configurations, with minimal differences between the original (FA) and translated (EN) data and only marginal variations due to template changes. 
Using English translation does not consistently improve the results. In the EA, translation to English has a minimal overall impact, with two models showing no change (Llama3-8B and Mixtral-7B). For Qwen2-7B, we observe a slight decrease in the English version of the data. The same trend is for the SA task, where all models have a lower average F1 score over English texts, except for the Qwen2-7B model, whose translation increases the average F1-score from 0.42 to 0.47, which is negligible. Regarding different prompt templates, we do not observe significant improvements over a specific template in the EA task. However, in the SA task, template (II) performs better than both the Farsi and English versions of the data, except for the Qwen2-7B model.
These findings suggest that both prompt design and data translation strategies for these subjective tasks in Farsi have a slight influence on model outcomes, particularly in EA.

\subsection{Experiment 2: LLM Evaluation}
Based on results in \Cref{tab:combined-macro-avg-comparison}, we use the prompt template (II) and datasets in Farsi (no translation) in a zero-shot setup to evaluate the different LLMs' performances across the selected datasets in \Cref{sec:datasets}. \Cref{tab:combined-analysis} presents the macro average F1-score, across all tasks, datasets, and models. Performance is benchmarked against two baselines: a random classifier and a Most Frequent Class (MFC) baseline. When we examine the performance of LLMs across all tasks, Qwen2-7B consistently outperforms the other models, achieving the highest average F1-Scores in EA (0.370), SA (0.563), and TD (0.809). As expected, Random and MFC baselines show lower scores than LLMs. 
At the dataset level, the Qwen2-7B shows higher scores over the ArmanEmo and LetHerLearn datasets. Over the EmoPars dataset, Llama3-8B achieves a 0.227 average F1-score, which is slightly better than Qwen2-7B's 0.218 average F1-score. We also report the results per label for each dataset and task in \Cref{tab:toxicity_f1,tab:emotion_f1_avg,tab:sentiment_f1_avg}.


\subsection{Experiment 3: Fine-Tuned LM Evaluation}
We fine-tune separate XLM-RoBERTa models on the train splits of each of the datasets and evaluate them on their corresponding test splits. The summary of the F1-score is presented in \Cref{tab:combined-analysis}. It demonstrates that XLM-RoBERTa performs overall better than LLMs and baseline models, except in the TD task for the Pars-off dataset. Although the model shows the strongest performance in the TD task (average F1 = 0.851), indicating its effectiveness in identifying toxic content. In SA, the model performs well on MirasOpinion and ParsABSA (average  F1 = 0.855), but its performance drops on SentiPars, suggesting inconsistencies across sentiment datasets. The EA task appears to be the most challenging task, with only moderate performance on ArmanEmo and LetHerLearn (average F1 = 0.641) and a lower score on the EmoPars dataset (F1 = 0.380). These findings highlight the impact of dataset characteristics on model performance and indicate the task difficulty for EA (complete result is available in the appendix in \Cref{tab:xlmr-finetune}).





\section{Conclusion}
Research on subjective tasks in the Farsi language has grown over the past five years, with a notable increase in the SA and TD research starting in early 2020. Most work has focused on two main data sources: social media data, such as tweets, and e-commerce data highlighting the challenge of data source scarcity in Farsi. Our review of over 110 papers identified several gaps, including a lack of diverse datasets, annotation information, and demographic features in subjective tasks, particularly for EA. 
These gaps include demographic disparities such as age and gender. Our experiments indicate that LLMs perform relatively poorly on EA tasks in Farsi but show stronger performance on SA and TD. Additionally, fine-tuning consistently improves performance across all tasks.

\section{Limitations and Ethical Considerations}
We acknowledge several limitations in our study. First, our evaluation relies heavily on existing publicly available datasets, which may not comprehensively capture the linguistic, cultural, or topical diversity of the Farsi language. These datasets may contain annotation biases, domain-specific skew, or inconsistencies that could affect model performance and generalizability.

\bibliography{anthology,custom}

\appendix
\section{Survey Analysis}


\section{Prompt Templates}
\label{appx:promting}

\subsection{Prompt Templates for EA}

\begin{itemize}
    \item \textbf{Template (I): } Given a text, identify the main emotion expressed. You have to pick one of the following seven emotions: sadness, hate, anger, happiness, fear, surprise, or other. Only answer with emotion and omit explanations. Emotion:


\item \textbf{Template (II): } You will be presented with a given comment sourced from X, Instagram, or Digikala. Pick one emotion from sadness, hate, anger, happiness, fear, surprise, or other that describes the emotion of the tweet or comment the best. Your response should only contain one of the emotions. No other output is allowed.
\end{itemize}

\subsection{Prompt Templates for SA}
\begin{itemize}
    \item \textbf{Template (I):} Given a text, identify the sentiment expressed. You have to pick one of the following three sentiments: positive, negative, neutral. Only answer with the sentiment and omit explanations. Sentiment: 

    
    \item \textbf{Template (II)} You will be presented with a comment from Digikala. Pick one sentiment from positive, negative, or neutral that describes the sentiment of the comment the best. Your response should only contain one of sentiment. No other output is allowed.

\end{itemize}

\subsection{Model hyperparameters}
\subsection{Models} \label{app:models}
Llama3 \cite{grattafiori2024llama} is an open-access collection of pre-trained and fine-tuned LLMs ranging in scale from 8 billion to 70 billion parameters and launched in September 2024. We examine Llama3-8B model. We use Qwen2-7B-Instruct model that published in November 2024 \cite{yang2024qwen2}. Mistral-7b is also an open-source LM launched in September 2023 \cite{jiang2024mixtral}. Among the models released by Mistral, we test Mixtral-8x7B-Instruct-v0.1, and we access these models via HuggingFace~\cite{wolf2019huggingface}. 

All responses were collected during July 2024 to March 2025. We run all our experiments on a server with three NVIDIA RTX A6000 and 48GB of RAM.

\textbf{XLM-RoBERTa} The hyperparameters
for the XLM-RoBERTa is three epochs,
batch size of 16, learning\_rate of 2e-5, optimizer
of Adam and the maximum length of 128.

\begin{table}[h]
\centering
\begin{adjustbox}{max width=\columnwidth}
\begin{tabular}{llcccc}
\toprule
\textbf{Task} & \textbf{Dataset} & \textbf{Acc.} & \textbf{Prec.} & \textbf{Rec.} & \textbf{F1} \\
\midrule
\multirow{3}{*}{\textbf{EA}} 
& ArmanEmo & 0.620 & 0.642 & 0.640 & 0.630 \\
& LetHerLearn & 0.644 & 0.670 & 0.649 & \textbf{0.653}  \\
& EmoPars & 0.480 & 0.421 & 0.428 & 0.380 \\
\cmidrule(lr){2-6}
& \textit{Avg.} & 0.581 & 0.578 & 0.572 & 0.554 \\
\midrule
\multirow{3}{*}{\textbf{SA}} 
& ParsABSA & 0.859 & 0.865 & 0.847 & \textbf{0.856} \\
& SentiPars  & 0.709 & 0.550 & 0.585 & 0.564 \\
& MirasOpinion & 0.874 & 0.856 & 0.851 & 0.854  \\
\cmidrule(lr){2-6}
& \textit{Avg.} & 0.814 & 0.757 & 0.761 & 0.758 \\
\midrule
\multirow{3}{*}{\textbf{TD}} 
& Phate  & 0.791 & 0.751 & 0.745 & 0.748 \\
& Pars-OFF  & 0.884 & 0.875 & 0.840 & 0.854 \\
& PHICAD  & 0.959 & 0.950 & 0.949 & \textbf{0.950} \\
\cmidrule(lr){2-6}
& \textit{Avg.} & 0.878 & 0.859 & 0.845 & 0.851 \\
\bottomrule
\end{tabular}
\end{adjustbox}
\caption{Performance of XLM-RoBERTa fine-tuned separately on nine datasets across three tasks. Each row reports Accuracy, Precision, Recall, and F1-score on the test set. The highest F1-score is highlighted in bold per dataset.}
\label{tab:xlmr-finetune}
\end{table}

\subsection{Emotion Analysis}
\label{appx:EA-analysis}
\Cref{tab:emotion_f1_avg} shows the performance of the LLMs across different emotions for each dataset.

\begin{table*}[h]
\centering

\begin{tabular}{lccccc}
\toprule
\textbf{Dataset} & \textbf{Emotion} & \textbf{Mixtral-7B} & \textbf{Llama3-8B} & \textbf{Qwen2-7B} & \textbf{Avg.} \\
\midrule
\multirow{7}{*}{Letherlearn} 
& Anger & 0.241 & 0.493 & 0.358 & 0.364 \\
& Disgust & 0.189 & 0.056 & 0.183 & 0.143 \\
& Fear & 0.488 & 0.461 & 0.458 & 0.469 \\
& Happiness & 0.423 & 0.545 & 0.560 & 0.509 \\
& Sadness & 0.447 & 0.503 & 0.511 & 0.487 \\
& Surprise & 0.420 & 0.264 & 0.268 & 0.317 \\
& Other & 0.228 & 0.080 & 0.345 & 0.218 \\
\midrule
\multirow{7}{*}{Arman} 
& Anger & 0.021 & 0.455 & 0.456 & 0.311 \\
& Fear & 0.761 & 0.760 & 0.733 & 0.751 \\
& Hate & 0.109 & 0.241 & 0.441 & 0.264 \\
& Happiness & 0.075 & 0.521 & 0.584 & 0.393 \\
& Sadness & 0.414 & 0.489 & 0.480 & 0.461 \\
& Surprise & 0.465 & 0.440 & 0.483 & 0.463 \\
& Other & 0.231 & 0.075 & 0.393 & 0.233 \\
\midrule

\multirow{6}{*}{EmoPars} 
& Anger & 0.262 & 0.307 & 0.220 & 0.263 \\
& Fear & 0.141 & 0.162 & 0.177 & 0.160 \\
& Hate & 0.014 & 0.046 & 0.154 & 0.071 \\
& Happiness & 0.247 & 0.294 & 0.300 & 0.280 \\
& Sadness & 0.288 & 0.240 & 0.256 & 0.261 \\
& Surprise & 0.173 & 0.066 & 0.202 & 0.147 \\
\bottomrule
\end{tabular}
\caption{F1 Scores for Emotion Analysis Across Datasets and Models with Average.}
\label{tab:emotion_f1_avg}
\end{table*}

\subsection{Sentiment Analysis}
\label{appx:SA-analysis}
\Cref{tab:sentiment_f1_avg} shows the performance of the LLMs across different sentiments for each dataset. Mixtral-7B and Llama3-8B can not capture ``very negative'' and ``very positive'' labels.

\begin{table*}[h]
\centering

\begin{tabular}{lccccc}
\toprule
\textbf{Dataset} & \textbf{Sentiment} & \textbf{Mixtral-7B} & \textbf{Llama3-8B} & \textbf{Qwen2-7B} & \textbf{Avg.} \\
\midrule
\multirow{3}{*}{MirasOpinion} 
& Negative & 0.619 & 0.631 & 0.656 & 0.635 \\
& Neutral & 0.138 & 0.498 & 0.592 & 0.409 \\
& Positive & 0.736 & 0.812 & 0.800 & 0.783 \\
\midrule
\multirow{3}{*}{Pars-ABSA} 
& Negative & 0.619 & 0.627 & 0.616 & 0.621 \\
& Neutral & 0.138 & 0.332 & 0.336 & 0.269 \\
& Positive & 0.736 & 0.741 & 0.734 & 0.737 \\
\midrule
\multirow{5}{*}{Sentipers} 
& Very Negative & 0.000 & 0.000 & 0.058 & 0.019 \\
& Negative & 0.560 & 0.563 & 0.570 & 0.564 \\
& Neutral & 0.675 & 0.593 & 0.664 & 0.644 \\
& Positive & 0.520 & 0.576 & 0.586 & 0.561 \\
& Very Positive & 0.000 & 0.620 & 0.341 & 0.320 \\
\bottomrule
\end{tabular}
\caption{F1 Scores for Sentiment Analysis Across Datasets and Models with Average.}
\label{tab:sentiment_f1_avg}
\end{table*}

\subsection{Toxicity Detection}
\label{appx:TD-analysis}
\Cref{tab:toxicity_f1} shows the performance of the LLMs across each dataset for detecting offensive/hate speech languagee.

\begin{table*}[htbp]
\centering
\begin{tabular}{lcccccc}
\toprule
\textbf{Dataset} & \textbf{Labels} & \textbf{Mixtral-7B} & \textbf{Llama3-8B} & \textbf{Qwen2-7B} & \textbf{Avg.} \\
\midrule
\multirow{2}{*}{Pars\_OFF} 
& not-offensive & 0.841 & 0.736 & 0.993 & 0.857 \\
& offensive & 0.640 & 0.656 & 0.857 & 0.718 \\
\midrule

\multirow{2}{*}{Phate} 
& not-hate & 0.692 & 0.553 & 0.720 & 0.655 \\
& hate & 0.673 & 0.778 & 0.409 & 0.620 \\
\midrule

\multirow{2}{*}{PHICAD} 
& not-hate & 0.911 & 0.887 & 0.990 & 0.929 \\
& hate & 0.667 & 0.653 & 0.894 & 0.738 \\

\bottomrule
\end{tabular}
\caption{Toxicity Detection F1 Scores Across Datasets and Models.}
\label{tab:toxicity_f1}
\end{table*}

\end{document}

%% file: tables/results_models_tasks.tex
\begin{table*}[ht]
\centering
\small
\renewcommand{\arraystretch}{1.2}
\begin{tabular}{ll|c|ccc|c|c}
\toprule
\textbf{Task} & \textbf{Dataset} & \textbf{XLM-RoB.} & \textbf{Llama3-8B} & \textbf{Mixtral-7B} & \textbf{Qwen2-7B} & \textbf{Random} & \textbf{MFC} \\
\midrule
\multirow{4}{*}{\textbf{EA}} 
& ArmanEmo & 0.630 & 0.426 & 0.296 & 0.510 & 0.135 & 0.061 \\
& LetHerLearn & 0.653 & 0.343 & 0.348 & 0.383 & 0.151 & 0.048 \\
& EmoPars & 0.380 & 0.227 & 0.188 & 0.218 & 0.152 & 0.048 \\
\cmidrule(lr){2-8}
& \textit{Avg.} & \textbf{0.554} & 0.332 & 0.277 & 0.370 & 0.146 & 0.052 \\
\midrule
\multirow{4}{*}{\textbf{SA}} 
& ParsABSA & 0.856 & 0.501 & 0.498 & 0.444 & 0.242 & 0.168 \\
& SentiPars & 0.564 & 0.453 & 0.351 & 0.562 & 0.199 & 0.108 \\
& MirasOpinion & 0.854 & 0.647 & 0.608 & 0.683 & 0.330 & 0.230 \\
\cmidrule(lr){2-8}
& \textit{Avg.} & \textbf{0.758} & 0.534 & 0.486 & 0.563 & 0.257 & 0.169 \\
\midrule

\multirow{4}{*}{\textbf{TD}} 
& Phate & 0.748 & 0.674 & 0.682 & 0.562 & 0.504 & 0.412 \\
& Pars-OFF & 0.854 & 0.696 & 0.741 & 0.925 & 0.491 & 0.412 \\
& PHICAD & 0.950 & 0.770 & 0.789 & 0.942 & 0.500 & 0.418 \\
\cmidrule(lr){2-8}
& \textit{Avg.} & \textbf{0.851} & 0.640 & 0.737 & 0.809 & 0.499 & 0.414 \\
\bottomrule
\end{tabular}
\caption{Macro Average F1-Scores for each model and dataset across three tasks: SA = Sentiment Analysis, TD = Toxicity Detection, EA = Emotion Analysis. Averages are calculated per task. \textbf{XLM-RoB.} is XLM-RoBERTa fine-tuned separately on nine datasets across three tasks. MFC is Most Frequent Class. The highest average F1-score is highlighted in bold per model.}
\label{tab:combined-analysis}
\end{table*}